\documentclass{article}
\usepackage{arxiv}
\usepackage[utf8]{inputenc}
\usepackage[T1]{fontenc} 
\usepackage{hyperref}  
\usepackage{url}    
\usepackage{booktabs}
\usepackage{amsfonts} 
\usepackage{nicefrac} 
\usepackage{microtype}
\usepackage{fancyhdr}
\usepackage{graphicx} 
\usepackage{subfig}
\usepackage{subcaption}
\usepackage{amsmath}
\usepackage{colortbl}
\usepackage{rotating}
\usepackage[table]{xcolor}
\usepackage{tikz}
\usepackage{float}
\usetikzlibrary{shapes.geometric, arrows.meta, positioning}

\graphicspath{{figures/}}     

\newlength{\subcolumnwidth}

\newcommand{\nextsubcolumn}[1][]{
  \cr\noalign{\hfill}
  \if\relax\detokenize{#1}\relax\else\hsize=#1\setlength{\subcolumnwidth}{\hsize}\fi
}

\title{KAN-SR: A Kolmogorov-Arnold Network Guided Symbolic Regression Framework
}

\author{\href{https://orcid.org/0009-0002-5323-5957}{\includegraphics[scale=0.06]{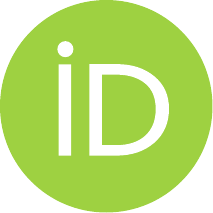} \textbf{Marco A. Bühler}}\\
  ETH Zürich, Switzerland\\
  \texttt{marco.buehler@chem.ethz.ch}
  \and 
  \href{https://orcid.org/0000-0001-6074-8473}{\includegraphics[scale=0.06]{orcid.pdf} \textbf{Gonzalo Guillén-Gosálbez}}\thanks{Corresponding Author} \\
  ETH Zürich, Switzerland\\
  \texttt{gonzalo.guillen.gosalbez@chem.ethz.ch}}

\begin{document}
\maketitle

\begin{abstract}
We introduce a novel symbolic regression framework, namely KAN-SR, built on Kolmogorov Arnold Networks (KANs) which follows a divide-and-conquer approach. Symbolic regression searches for mathematical equations that best fit a given dataset and is commonly solved with genetic programming approaches. We show that by using deep learning techniques, more specific KANs, and combining them with simplification strategies such as translational symmetries and separabilities, we are able to recover ground-truth equations of the Feynman Symbolic Regression for Scientific Discovery (SRSD) dataset. Additionally, we show that by combining the proposed framework with neural controlled differential equations, we are able to model the dynamics of an in-silico bioprocess system precisely, opening the door for the dynamic modeling of other engineering systems.
\end{abstract}


\section{Introduction}

Symbolic regression (SR) aims to recover closed‐form mathematical expressions that accurately model a given dataset. Unlike traditional regression approaches, which assume a fixed model structure, SR jointly infers both the functional form (equation) and its parameters (constants) from the data, producing models that are not only predictive, but also (to some extent) interpretable. This flexibility makes SR especially appealing in scientific domains, where understanding the underlying mechanisms can be more important than raw predictive performance~\cite{Brunton2015, Forster2024}.

Because of this, SR emphasizes simplicity and interpretability alongside predictive performance. The goal is to find concise mathematical relationships, ideally those that offer insight into the system being modeled~\cite{Lacava2021}. This contrasts with black-box machine learning models, which may perform well but often lack transparency and might extrapolate poorly.

Classical SR methods, such as genetic programming (GP)~\cite{Koza1994, Cranmer2023, Guimer2020, Schmidt2009}, explore the space of symbolic expressions using evolutionary algorithms applied to expression trees~\cite{jin2020}. Although flexible, GP-based methods tend to be sample‐inefficient, computationally expensive, and often face difficulties with scalability. More recent approaches have tried to overcome these limitations using pre‐trained language models~\cite{Biggio2021, Grayeli2024, Kamienny2022}, reinforcement-learning~\cite{Petersen2019}, hybrid frameworks that integrate deep learning with GP‐style symbolic search~\cite{Mundhenk2021, Landajuela2022} and physics-inspired heuristics~\cite{Udrescu2020}.

Despite this progress, challenges remain. Many neural-symbolic systems still rely on discrete symbolic representations that are difficult to train end-to-end, limiting their ability to learn efficiently or produce more interpretable results. Furthermore, there is no universally adopted benchmark, which makes comparing SR methods inconsistent. A step toward covering this gap was the Symbolic Regression for Scientific Discovery (SRSD) benchmark by Matsubara \textit{et al.}~\cite{Matsubara2022}, which introduced 240 datasets inspired by the Feynman lectures, with sampling which corresponds to the real measured phenomena and the inclusion of irrelevant variables to test model robustness.

In engineering sciences such as chemical engineering and bioengineering, the insights created by the SR algorithm can lead to faster and more robust process development or a deeper understanding of phenomena. Applications range from finding kinetic models to reactions~\cite{Forster2024, Cohen2024, Wilson2017, Servia2023}, designing new catalysts~\cite{Weng2020} and describing different transport phenomena~\cite{Ansari2022}. Additionally, dynamic systems are at the core of many of these applications, and identifying these governing equations in a dynamic environment is an open problem. An algorithm that is tailored to identify differential equations is SINDy~\cite{Brunton2015} which leverages sparse regression to find the best fitting combination of hand-made equation blocks.

In this paper, we introduce \textbf{KAN-SR}, a symbolic regression framework based on Kolmogorov-Arnold Networks (KANs)~\cite{Liu2024, Liu2024b}. KANs are a recent neural network architecture inspired by the Kolmogorov-Arnold representation theorem. Instead of traditional neurons with fixed activation functions and linear weights, KANs use learnable univariate functions, allowing more expressive and often more interpretable function approximations.

To adapt KANs for symbolic regression, we develop a hybrid extraction pipeline that combines differentiable KAN training with symbolic simplification strategies inspired by AI Feynman~\cite{Udrescu2020}. By applying these symbolic simplifications, e.g. symmetries and separabilities, the initial complex problem can be decomposed into simpler subproblems, which can often be directly fitted by a single-layer KAN. After fitting a KAN to the data, we decompose the network into modular algebraic components which are then matched to a symbolic library via non-linear least squares. This enables the extraction of compact, interpretable expressions without relying on brute‐force search or discrete enumeration.

We evaluate KAN‐SR on the SRSD dataset, comparing it against state‐of‐the‐art symbolic regression models evaluated in Matsubara \textit{et al.}~\cite{Matsubara2022}. Our results show that KAN‐SR achieves competitive or superior performance in symbolic recovery, particularly in scenarios involving compositional structure, extrapolation, and noisy inputs with irrelevant variables.

Furthermore, we extend KAN‐SR to an in-silico dynamic biological system, demonstrating its application to a bioprocess modeling case study, where it recovers interpretable kinetic rate equations~\cite{Forster2024}. Overall, our work opens up new avenues for the application of KANs to SR problems in a range of engineering problems, including the modeling of dyamic systems.

\subsection*{Contributions}

\begin{itemize}
    \item We propose a symbolic regression framework based on Kolmogorov‐Arnold Networks and symbolic modeling, unifying compositional deep learning and symbolic extraction.
    \item We introduce a symbolic extraction and simplification pipeline inspired by \textit{AI Feynman}~\cite{Udrescu2020}, using a comprehensive and extendable univariate function library in combination with KANs.
    \item We evaluate our approach on the SRSD benchmark~\cite{Matsubara2022}, showing improvements in symbolic recovery.
    \item We demonstrate the extension of our method to dynamic systems through a bioprocess modeling case study, leveraging neural-controlled differential equations.
\end{itemize}

\section{Methods}
SR searches for an equation \(y(x)=\Theta(x,\theta)\) in the space of expressions \(\Theta\) and constants \(\theta\), to find the best fit combination that minimizes an objective function, which often is a combination of the fitting error and a complexity measure of the equation, such as the Bayesian Information Criterion (BIC). SR assumes that there is an analytical ground truth of the form \(y(x)=\Theta^*(x,\theta^*) + \epsilon\) that created the training data \(x\in \mathbb{R}^d\) with the target \(y\in \mathbb{R}\) in the presence of Gaussian noise \(\epsilon\). 

\paragraph{Implementation details:}
KAN-SR was implemented in Python using Jax~\cite{jax2018github} as the base for all calculations, Equinox~\cite{kidger2021equinox} for the KANs, Optimistix~\cite{optimistix2024} for the non-linear least squares regression and optimizers, Optax~\cite{deepmind2020jax} for first-order optimizers and Diffrax~\cite{kidger2021on} for the neural differential equations~\cite{chen2019}.

\subsection{Workflow}

The proposed symbolic regression framework is highly flexible and involves several hyperparameters and algorithmic components. This flexible design is motivated by the fact that there is no universal KAN architecture that can accommodate all target functions, even when strong regularization is applied. Consequently, the workflow consists of a sequence of optional and configurable steps, some of which can be enabled or skipped, depending on the problem characteristics and performance criteria.

The general workflow is as follows, see Figures \ref{fig:mainworkflow} for a more detailed illustration:

\begin{enumerate}
    \item \textbf{Preprocessing:} Normalize or scale the input variables to improve the numerical stability and convergence.
    \item \textbf{Brute-force search:} Optionally perform an exhaustive search over simple expressions.
    \item \textbf{Single-layer KANs with a single unit:} Attempt a symbolic approximation using minimal architectures with a summation or multiplication unit.
    \item \textbf{Single-layer KANs with multiple units:} Expand the search space by allowing multiple units within a single-layer architecture.
    \item \textbf{Simplification and subproblem decomposition:} Optionally identify structural simplifications or separable components in the learned expression, and recursively restart the procedure on the subproblems.
    \item \textbf{Output transformation:} Optionally apply transformations to the output (e.g., logarithmic, exponential) to simplify the functional form and rerun the symbolic search on the transformed target.
    \item \textbf{Deep KAN fitting:} As a final step, if chosen, employ deeper KAN architectures to capture more complex or hierarchical functional relationships if simpler models fail, see Figure \ref{fig:deepkan} for an illustration.
\end{enumerate}

\begin{figure}[htbp]
    \centering
    \large
     \begin{minipage}[t]{0.48\textwidth}
        \centering
    \resizebox{0.9\textwidth}{!}{
    \begin{tikzpicture}[
        node distance=1.2cm and 2.5cm,
        every node/.style={align=center},
        decision/.style={diamond, draw, aspect=2, inner sep=1pt},
        process/.style={rectangle, draw, rounded corners},
        optional/.style={rectangle, draw, rounded corners, dashed},
        line/.style={->, thick},
        scale=1, transform shape
    ]

    \node (input) [process] {Input Data};
    \node (preprocess) [process, below of=input] {Preprocessing};
    \node (brute) [optional, below of=preprocess] {Brute-force Symbolic Matching};
    \node (match0) [decision, below of=brute, yshift=-0.5cm] {error $\leq$ threshold};
    \node (return0) [process, right of=match0, xshift=4.5cm] {Return Symbolic Model};

    \node (single) [process, below of=match0, yshift=-1cm] {Single-Unit KAN};
    \node (extract1) [process, below of=single] {Symbolic Extraction};
    \node (match1) [decision, below of=extract1, yshift=-0.5cm] {error $\leq$ threshold};
    \node (return1) [process, right of=match1, xshift=4.5cm] {Return Symbolic Model};

    \node (multi) [process, below of=match1, yshift=-1cm] {Multi-Unit KAN};
    \node (extract2) [process, below of=multi] {Symbolic Extraction};
    \node (match2) [decision, below of=extract2, yshift=-0.5cm] {error $\leq$ threshold};
    \node (return2) [process, right of=match2, xshift=4.5cm] {Return Symbolic Model};

    \node (simplify) [optional, below of=match2, yshift=-1cm] {Simplification / Transformation};
    \node (check_simp) [decision, below of=simplify, yshift=-0.75cm] {Simplification Found};
    \node (simplinp) [process, left of=check_simp, xshift=-4cm] {Simplified Input};
    \node (DeepKan) [optional, right of=check_simp, xshift=4.75cm] {Deep KAN Loop, Figure~\ref{fig:deepkan}};

    \draw [line] (input) -- (preprocess);
    \draw [line] (preprocess) -- (brute);
    \draw [line] (brute) -- (match0);
    \draw [line] (match0) -- node[above] {Yes} (return0);
    \draw [line] (match0) -- node[left] {No} (single);

    \draw [line] (single) -- (extract1);
    \draw [line] (extract1) -- (match1);
    \draw [line] (match1) -- node[above] {Yes} (return1);
    \draw [line] (match1) -- node[left] {No} (multi);

    \draw [line] (multi) -- (extract2);
    \draw [line] (extract2) -- (match2);
    \draw [line] (match2) -- node[above] {Yes} (return2);
    \draw [line] (match2) -- node[left] {No} (simplify);

    \draw [line] (simplify) -- (check_simp);
    \draw [line] (check_simp) -- node[above] {Yes} (simplinp);
    \draw [line] (simplinp.north) |- (input.west);
    
    \draw [line] (check_simp) -- node[above] {No} (DeepKan);

    \node at ([xshift=2.4cm, yshift=2.2cm] brute.east) [optional, rotate=0] {\textit{Optional Components}};

    \end{tikzpicture}
    }
    \caption{Main symbolic regression workflow. Optional modules include brute-force matching, simplification, and deep KAN fitting. The pipeline returns early if a symbolic match is found, otherwise, it proceeds through increasingly expressive modeling stages. If the threshold is not reached after a full completion of the algorithm, the best equation found during the search is returned.}
    \label{fig:mainworkflow}
    \end{minipage}
    \hfill
    \begin{minipage}[t]{0.48\textwidth}
        \centering
        \resizebox{0.75\textwidth}{!}{
    \begin{tikzpicture}[
        node distance=1.3cm and 3.2cm,
        every node/.style={align=center},
        decision/.style={diamond, draw, aspect=2, inner sep=1pt},
        process/.style={rectangle, draw, rounded corners},
        line/.style={->, thick},
        scale=1, transform shape
    ]

    \node (start) [process] {Start Deep KAN Loop};
    \node (depth_init) [process, below of=start] {Set $d = 1$};
    \node (depthcheck) [decision, below of=depth_init] {$d > d_{\text{max}}$};
    \node (fail) [process, right of=depthcheck, xshift=3cm] {Return Best So Far};

    \node (width_init) [process, below of=depthcheck, yshift=-0.5cm] {Set $w = w_{\text{min}}$};
    \node (fit) [process, below of=width_init, yshift=-0.25cm] {Fit KAN with $(d, w)$};
    \node (extract) [process, below of=fit, yshift=-0.25cm] {Symbolic Extraction};
    \node (match) [decision, below of=extract, yshift=-0.2cm] {error $\leq$ threshold};
    \node (return) [process, right of=match, xshift=4cm] {Return Symbolic Model};

    \node (widthcheck) [decision, below of=match, yshift=-1cm] {$w < w_{\text{max}}$};
    \node (incwidth) [process, left of=widthcheck, xshift=-1.75cm] {Increment $w$};

    \node (incdepth) [process, below of=widthcheck, yshift=-0.3cm] {Increment $d$};

    \draw [line] (start) -- (depth_init);
    \draw [line] (depth_init) -- (depthcheck);
    \draw [line] (depthcheck) -- node[above] {Yes} (fail);
    \draw [line] (depthcheck) -- node[left] {No} (width_init);
    \draw [line] (width_init) -- (fit);
    \draw [line] (fit) -- (extract);
    \draw [line] (extract) -- (match);
    \draw [line] (match) -- node[above] {Yes} (return);
    \draw [line] (match) -- node[left] {No} (widthcheck);
    \draw [line] (widthcheck) -- node[right] {No} (incdepth);
    \draw [line] (widthcheck) -- node[above] {Yes} (incwidth);
    \draw [line] (incwidth.north) |- (fit);
    \path (incdepth.west) -- ++(-3.25,0) coordinate (pullLeft2);
    \draw [line] (incdepth.west) |- (pullLeft2) |- (depthcheck);

    \end{tikzpicture}
    }
    \caption{Nested deep KAN fitting loop. For each depth $d$, the system explores increasing widths $w$ up to a maximum. If no symbolic match is found, the depth is incremented until the limit is reached.}
    \label{fig:deepkan}
    \end{minipage}
\end{figure}

\subsection{Brute-Force Symbolic Matching}
To rapidly identify simple and low-complexity symbolic relationships, we implement a brute-force symbolic regression module as a first step in the workflow. This stage exhaustively fits a set of simple multivariate multiplicative expressions to the data.

The expression library includes forms such as:
\[
f(x) = \theta_1 x_0 x_1 + \theta_2, \quad
f(x) = \frac{\theta_1 x_0}{x_1 + \theta_2} + \theta_3, \quad
\]
as well as higher-order expressions such as: \[ f(x) = \theta_1 x_0 x_1 x_2 x_3 + \theta_2, \quad f(x) = \frac{\theta_1}{x_0 x_1 x_2 x_3 + \theta_2} + \theta_3. \]

Each individual expression is defined as a parameterized function \( f(x; \theta) \) and is fitted using a BFGS (Broyden--Fletcher--Goldfarb--Shanno) algorithm. If the resulting error of the best fitting equation is below a threshold, the symbolic form is accepted without further modeling.

This brute-force module is especially effective when the target function has a simple algebraic form and allows for a much faster symbolic identification before applying more expensive models.

\subsection{Kolmogorov--Arnold Networks}
In the original KAN formulation~\cite{Liu2024, Liu2024b}, univariate activation functions are represented using B-splines. Although expressive, this implementation has practical limitations: the grid of spline knots can be manually extended during training allowing for better accuracy but this stops training and an additional fit has to be performed. Additionally the evaluation of the splines is slow and can thus become a computational bottleneck. To address these issues, we adopt the improved Fast-KAN approach~\cite{Li2024}, which replaces B-splines with radial basis functions for greater efficiency.

In our work, the learnable univariate activation is implemented using a \textit{Reflectional Switch Activation Function} (RSWAF)~\cite{Athanasios2024}, defined as:

\[
    w_i \left( s_i - \tanh^2\left( \frac{x - c_i}{h_i} \right) \right),
\]

where:
\begin{itemize}
    \item \(w_i\) is a learnable weight that modulates the contribution of the \(i\)-th basis function,
    \item \(s_i\) is the reference activation level,
    \item \(c_i\) denotes the center of the activation,
    \item \(h_i\) is a scale parameter that controls the sharpness of the activation around \(c_i\).
\end{itemize}

This formulation yields smooth, localized basis functions centered on \( c_i \), and defined in the complete domain. The compositional flexibility of these units allows the network to approximate a wide range of non-linear univariate functions.

\begin{figure}[htbp]
    \centering
    \subfloat[Random initialization of five basis functions with a uniform grid.]{
        \includegraphics[width=0.32\textwidth]{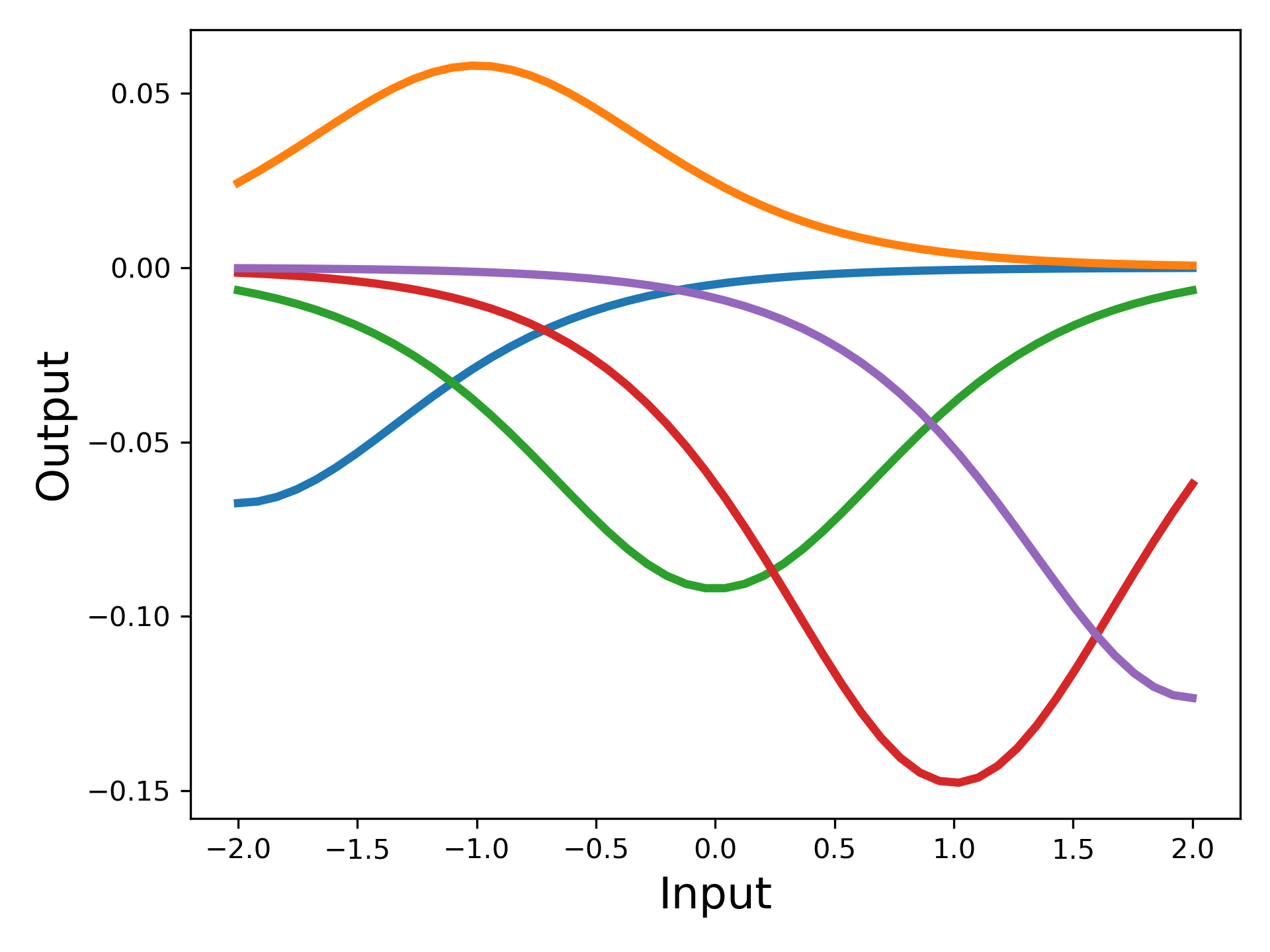}
    }
    \hfill
    \subfloat[Basis functions after training.]{
        \includegraphics[width=0.32\textwidth]{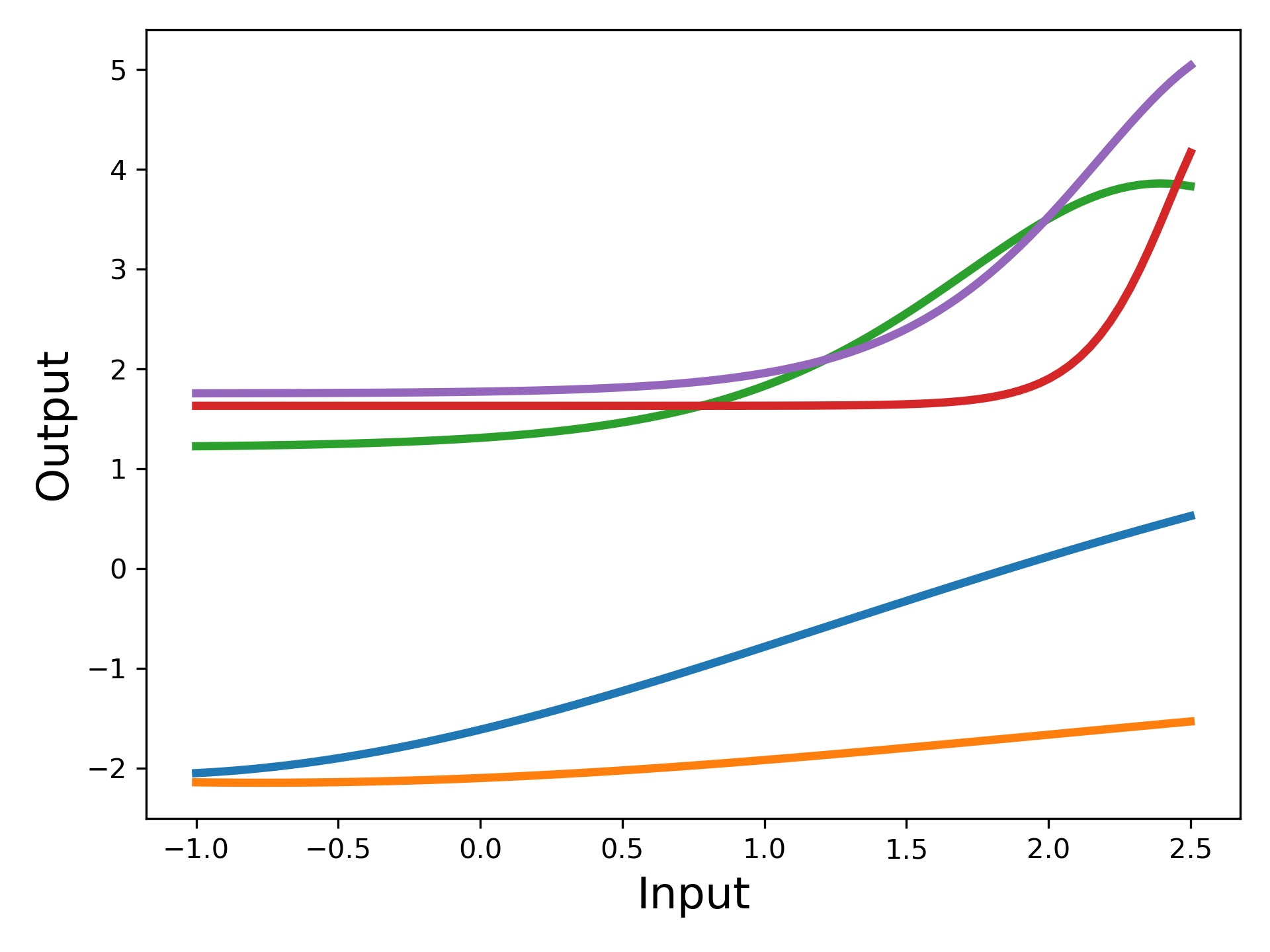}
    }
    \hfill
    \subfloat[Element-wise sum over all five basis functions of (b); Equation \ref{eq:final_activation}.]{
        \includegraphics[width=0.32\textwidth]{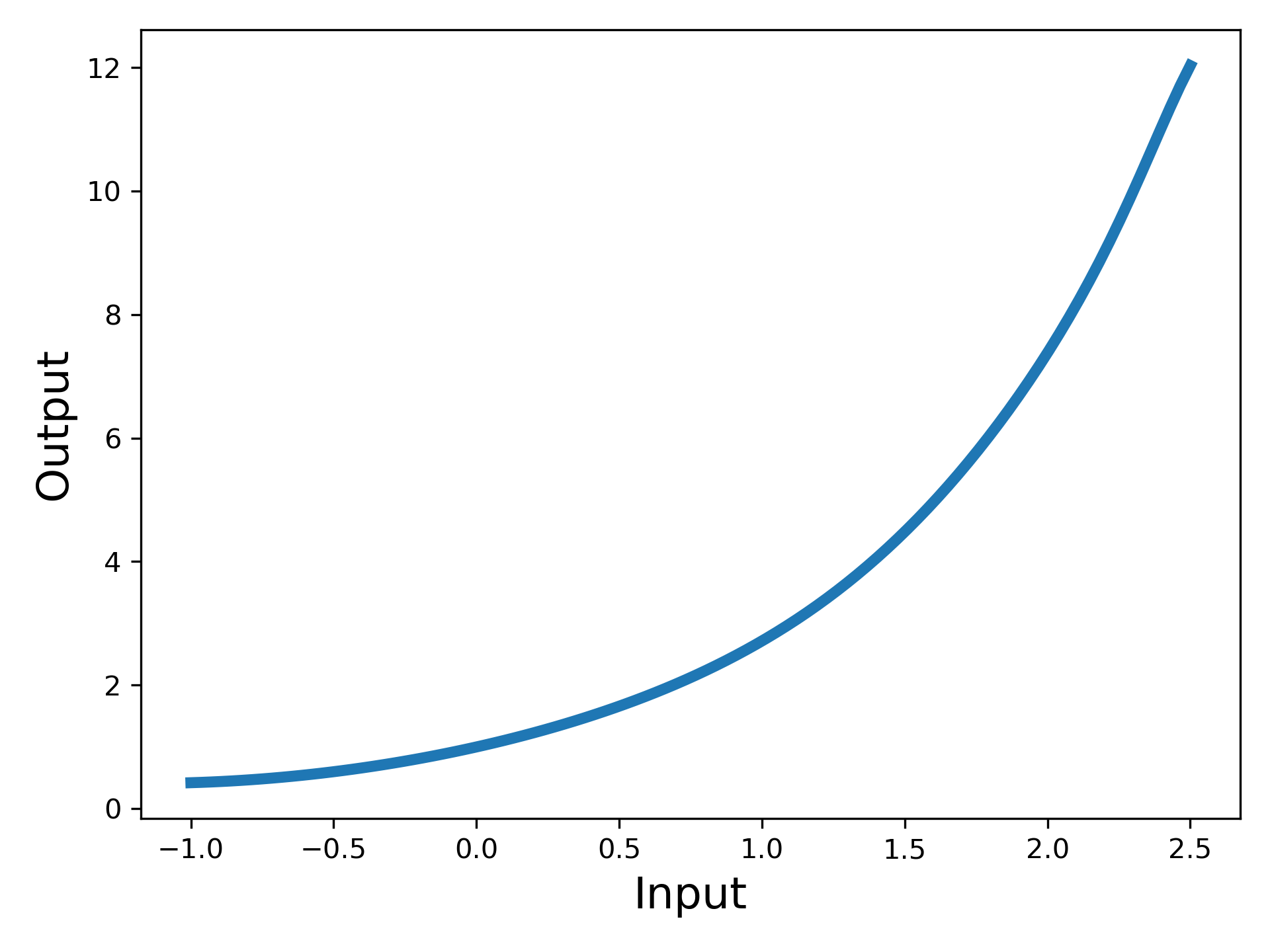}
    }
    \caption{Example of the training of a single univariate activation function, where the grid is directly adapted during training, to best fit the target exponential function. The activation function is only evaluated over the input range of the variable to extract the best fitting equation.}
    \label{fig:rbf_trained}
\end{figure}

The final activation function is then of the form:
\begin{align}
    \phi(x) = \sum_i w_i \left( s_i - \tanh^2\left( \frac{x - c_i}{h_i} \right) \right)
    \label{eq:final_activation}
\end{align}
This reformulation from B-Splines to radial basis functions allows for faster training, decreasing the time needed for the forward and backward pass. Empirically, we find that most target functions can be approximated with just five such basis functions per univariate activation, resulting in 20 learnable parameters per activation, the only limitation being high-frequency periodic signals. By adapting these weights via gradient descent, the KAN can learn a variety of univariate functions (see Figure \ref{fig:rbf_trained} for an example).

However, a limitation of the original KAN is its inherent inability to directly represent multiplicative interactions of the form
\[
f(x_1, x_2) = g(x_1) \cdot h(x_2)
\]
using a single-layer architecture. Such factorizable functions require multiple layers under a purely additive composition due to the lack of interaction terms.

To overcome this, we incorporate ideas from the multiplicative KAN extension~\cite{Liu2024b} and explicitly allow additive and multiplicative combinations of learnable activations. Specifically, instead of computing only additive outputs, we also construct terms of the form:
\[
f(x_1, \dots, x_n) = \prod_{i=1}^n \phi_i(x_i),
\]
where each \( \phi_i(x_i) \) is the learnable univariate function parameterized as above. This allows the model to directly represent separable multiplicative functions common in scientific domains.

Our final model architecture of a single-layer KAN consists of a sum over \( K \) such sub-networks (or units), where each unit computes a composition over the full input space via either summation or multiplication:

\[
f(x_1, \dots, x_n) = \sum_{k=1}^{K} \mathcal{C}_k\left( \phi_{k1}(x_1), \phi_{k2}(x_2), \dots, \phi_{kn}(x_n) \right),
\]
\[
\mathcal{C}_k(z_1, \dots, z_n) =
\begin{cases}
\sum_{i=1}^{n} z_i, & \text{if unit } k \text{ is additive}, \\
\prod_{i=1}^{n} z_i, & \text{if unit } k \text{ is multiplicative}.
\end{cases}
\]

This hybrid composition framework allows the network to represent a rich class of functions by flexibly combining both additive and multiplicative structures over the full input domain.

To apply the method to deeper or multi-output KANs, we extend the above equations to an additional dimension \(j\) that represents either the number of input features of the next layer or multiple output features.
\[
f_j(x_1, \dots, x_n) = \sum_{k=1}^{K_j} \mathcal{C}_k\left( \phi_{jk1}(x_1), \phi_{jk2}(x_2), \dots, \phi_{jkn}(x_n) \right),
\]

To guide learning, a base linear layer is included, as in the original implementation~\cite{Liu2024} with a GELU~\cite{hendrycks2023gaussianerrorlinearunits} activation function. This basis function is similar to residual connections, such that the final output is the sum of the basis function and the learnable activation functions.

\subsubsection*{Regularization}

Our final KAN architecture consists of multiple subunits, each operating on the full input space. Due to this, the model is often overspecified, whereas most symbolic expressions, particularly in scientific domains, often only depend on a small set of operators, each with a subset of input variables. To promote sparsity and thus interpretability, we introduce a regularization scheme designed to penalize unnecessary input usage and enforce compact functional representations, similar to the one proposed in the original implementation~\cite{Liu2024}, as described below.

For each subunit $k\in K$:
We construct a matrix \( A \in \mathbb{R}^{m \times n} \), where \( A_{ji} \) represents the mean absolute activation of \( \phi_{ji}(x_i) \) for input \( x_i\) \(i\in n\) and the output dimension $j\in m$. That is,
\[
A_{ji} = \left| \phi_{ji}(x_i) \right|
\]
We interpret \( A \) as capturing the degree to which each input dimension contributes to the output between samples.

By summing over the matrix we get the first penalty (magnitude):
\[
\mathcal{L}_{\text{magnitude}} = \sum_{j=1}^{m} \sum_{i=1}^{n} \left\| \phi_{ji}(x_i) \right\|_1
\]

To encourage sparsity on the whole structure, we again take matrix \(A_{ij}\) and use it to compute the row-wise and column-wise entropies (\(\mathcal{H}_{\text{row}}, \mathcal{H}_{\text{col}}\)), which are a measure of the diffuseness of activations across inputs and samples, respectively. Let:

\[
p^{\text{col}}_{ji} = \frac{A_{ji}}{\sum_{j=1}^{m} A_{ji} + \epsilon}, \quad 
p^{\text{row}}_{ji} = \frac{A_{ji}}{\sum_{i=1}^{n} A_{ji} + \epsilon}
\]

Then the entropy loss is defined as follows:
\[
\mathcal{L}_{\text{entropy}} = \mathcal{H}_{\text{col}}(A) + \mathcal{H}_{\text{row}}(A),
\]
\[
\mathcal{H}_{\text{col}}(A) = - \frac{1}{n} \sum_{i=1}^{n} \sum_{j=1}^{m} p^{\text{col}}_{ji} \log(p^{\text{col}}_{ji} + \epsilon)
\]
\[
\mathcal{H}_{\text{row}}(A) = - \frac{1}{m} \sum_{j=1}^{m} \sum_{i=1}^{n} p^{\text{row}}_{ji} \log(p^{\text{row}}_{ji} + \epsilon),
\]

These entropy terms encourage each subunit to focus on a small, confident set of inputs, reducing ambiguous or diffuse activation patterns. This is important to align the learned representations with the sparse structure of symbolic equations.

Finally, we include, when applicable, the \( \ell_1 \) penalty on the weights (\(W_{base}\)) of the base linear layer. The total regularization objective is the following, with subunits $k\in K$:

\[
\mathcal{L}_{\text{reg}}=\sum_k^K\mathcal{L}_{\text{reg},k} = \sum_k^K\lambda_0 (\lambda_1 \mathcal{L}_{\text{magnitude},k} + \lambda_2 \mathcal{L}_{\text{entropy},k} + \lambda_3 \left\| W_{\text{base},k} \right\|_1),
\]
where $\lambda_i$ are hyperparameters.

\subsection{Symbolic Simplification via Separability and Symmetry Detection}

Before training single-layer KANs, we apply the recursive decomposition strategies proposed by Udrescu et al.~\cite{Udrescu2020}. For this, a lightweight interpolating network is used to analyze the target function and detect structural properties, such as variable separability and symmetry, that can reduce the complexity of the problem. All the methods in this section come from Udrescu et al.~\cite{Udrescu2020} and only a short summary of the methods used is given here.

\paragraph{Separability Detection.}
We test whether the target function is approximately \textit{additively} or \textit{multiplicatively separable} across disjoint subsets of input variables. For each candidate split \( (S, \bar{S}) \), we define:

\begin{itemize}
    \item \( f_S(x) \): with \( \bar{S} \) fixed to its mean,
    \item \( f_{\bar{S}}(x) \): with \( S \) fixed to its mean,
    \item \( f_0(x) \): with both subsets fixed to their mean.
\end{itemize}

We then compute normalized approximation errors:
\[
\epsilon_{\text{add}} = \frac{1}{\epsilon_{\text{val}}} \left\| y - \left(f_S(x) + f_{\bar{S}}(x) - f_0(x)\right) \right\|_1, \quad
\epsilon_{\text{mult}} = \frac{1}{\epsilon_{\text{val}}} \left\| y - \left(\frac{f_S(x) \cdot f_{\bar{S}}(x)}{f_0(x)}\right) \right\|_1,
\]
where \( \epsilon_{\text{val}} \) is the validation error of the interpolating network. If either error falls below a fixed threshold (\( \epsilon < 10 \)), the function is considered separable. The best performing subset is used further if more than one subset was found during the search. Then two new data sets are created that only depend on the individual subsets found, using the trained interpolating network to create the new targets y' and y''. This allows KANs to model lower-dimensional subproblems.

\paragraph{Symmetry Detection.}
We also test for input symmetries by applying structured transformations to input pairs \( (x_i, x_j) \) and measuring the invariance of the output. The following transformations are considered, with \( a \) a constant:

\begin{itemize}
    \item \textbf{Negative translation:} \( f(x_i + a, x_j + a)\approx f(x_i - x_j) \)
    \item \textbf{Positive translation:} \( f(x_i + a, x_j - a) \approx f(x_i + x_j) \)
    \item \textbf{Multiplicative:} \( f(a x_i, x_j / a) \approx f(x_i \cdot x_j) \)
    \item \textbf{Divisive:} \( f(a x_i, a x_j) \approx f(x_i / x_j) \)
\end{itemize}

Each transformation is applied and the transformed output is compared with the original (\(f(x_i, x_j)\) ) normalized by the validation error. If the relative error falls below a threshold, the symmetry is deemed valid.

When multiple symmetries or separabilities are detected, we apply the one with the lowest relative error. Symmetries are evaluated first because they reduce dimensionality without relying on model-generated targets. Separabilities depend on the network output and may introduce an additional approximation error.

These symbolic simplification steps reduce complexity and dimensionality, allowing KANs to focus on simpler subproblems, ultimately improving interpretability and performance on scientific regression tasks.

\subsection{Symbolic Expression Extraction}

To convert trained KANs into closed-form symbolic models, we implement a post-hoc symbolic extraction pipeline that approximates each univariate activation with interpretable expressions from a predefined, extendable function library.

Each univariate activation \( \phi_{ki}(x_i) \) within the trained KAN is evaluated over its effective input range, and a set of candidate symbolic functions are fitted minimizing the mean squared error (MSE). The candidate functions are drawn from a predefined library comprising polynomials, trigonometric forms, exponential and logarithmic functions, and domain-specific functions such as Michaelis-Menten kinetics (often used to model biological systems).

The best-fitting function is selected according to one of two criteria:
\begin{itemize}
    \item \textbf{Best fit:} The candidate with the lowest MSE.
    \item \textbf{Score-based:} A simple trade-off metric between MSE and expression complexity, penalizing longer and more complex expressions. The complexity is penalized according to the following formula: With \(\Theta\) being the complexity manually chosen of the fitted function.

    \[
    \text{Score} = \text{MSE} \cdot 5^{\Theta}
    \]

    The parameters \(\Theta\) for each equation are chosen arbitrarily and tuned during experimentation, with lower complexity equations (such as linear or constants) having the lowest values and complex equations (such as \(x\cdot\sin x\)) higher values. This scoring approach offers a simple, yet effective, trade-off between accuracy and interpretability. While it can be easily extended to incorporate more complex model selection criteria, such as the Akaike Information Criterion (AIC) or Bayesian Information Criterion (BIC), which balance likelihood with model complexity, we find that this lightweight heuristic performs well enough in practice. Moreover, multi-objective optimization could also be used to handle the trade-off between complexity and accuracy, yet this would also lead to more complex formulations. 
\end{itemize}

Once all symbolic approximations are determined, the network structure is symbolically propagated layer by layer. In the case of additive or multiplicative compositions (as specified by each sub-unit), symbolic expressions are either summed or multiplied accordingly. Intermediate expressions are simplified using SymPy~\cite{sympy} and constants are re-optimized using a BFGS algorithm to minimize final output error.

This procedure results in a fully symbolic approximation of the trained KAN model:
\[
f(x_1, \dots, x_n) \approx \sum_{k=1}^{K} \mathcal{C}_k\left( \hat{\phi}_{k1}(x_1), \dots, \hat{\phi}_{kn}(x_n) \right),
\]
where \( \hat{\phi}_{ki} \) are the symbolic approximations extracted for each activation and \( \mathcal{C}_k \in \{\text{sum}, \text{product}\} \) denotes the unit-level composition type.

\subsection{Output Transformation for Simplification}

If after simplifying and fitting various single-layer KANs, a good fitting function has not been found, we start by transforming the target variable \(y\). Target functions may include non-linear or nested structures that are difficult to simplify directly or require deeper KAN architectures. To address this, we apply transformations to the output space to make the function simpler for symbolic approximation.

Let \( f(x) \) denote the original target function. We apply a transformation \( T \), such that:
\[
T(f(x))
\]
has a simplified structure.

For example, if \(f(x_1, x_2) = \exp(\sin(x_1) + \sin(x_2)),\) then applying the logarithm yields 
\(\ln(f(x_1, x_2)) = \sin(x_1) + \sin(x_2),\)
which is significantly simpler. After identifying a symbolic approximation \( \hat{f} \) of the transformed function, we recover the original via the inverse transformation:
\(
f(x_1, x_2) = \exp(\hat{f}(x_1, x_2)).
\)

\subsection{Neural Controlled Differential Equations for Noise-Robust Derivative Estimation}
\label{sec:NCDE}
Most SR algorithms are built and applied to static data. However, many engineering systems are inherently dynamic, making the direct application of these standard SR algorithms to them challenging. SINDy~\cite{Brunton2015} and ARGOS~\cite{Egan2024} are a purpose-built SR algorithm to discover the equations governing differential equations, but often require hand-made equation building blocks and sparse regression to guide the search. A different approach is ODEFormer~\cite{dascoli2023} which is a transformer-based architecture which directly predicts the governing equations from data profiles. 
Here, we discuss the use of our algorithm to model dynamic data, with emphasis on kinetic modeling of a bioprocess. In the modeling of dynamic systems, accurate derivative estimation is a crucial prerequisite to rediscover the underlying equations that govern them, and classical numerical differentiation methods (e.g., finite differences) are highly sensitive to noise, often producing unreliable gradients when applied to real-world, noisy measurements. To address this limitation, we use Neural Controlled Differential Equations (Neural CDEs)~\cite{kidger2020NCDE, kidger2021on} as a data-driven framework to extract smooth derivative estimates from noisy time series that will be used as input to KAN-SR.

Neural CDEs generalize recurrent neural networks (RNNs) by modeling the evolution of a hidden state in continuous time, where the input is treated as a control signal driving a differential equation. This idea extends the modeling paradigm introduced in Neural ODEs~\cite{chen2019}, whilst offering a more flexible framework for handling continuously arriving and irregularly sampled time series data.

A neural CDE models the evolution of a latent hidden state \( Z(t) \in \mathbb{R}^d \), driven by a known continuous control path \( U(t) \in \mathbb{R}^m \). The control path \( U(t) \) is assumed to be a smooth function of time, contrary to our observed state variables \( X(t) \in \mathbb{R}^n \), which can be noisy or discretely observed. It is typically constructed via interpolation from observational data or control inputs.

The dynamics follow:
\[\frac{dZ(t)}{dt} = f_\theta(Z(t)) \cdot \frac{dU(t)}{dt},\]
where \( f_\theta: \mathbb{R}^d \to \mathbb{R}^{d \times m} \) is a neural network that parameterizes the vector field and \( \frac{dU(t)}{dt} \) is the derivative of the control path that we can compute from the interpolation. Using the control path, the model can effectively steer the evolution of the hidden state to capture complex dependencies that are difficult to extract solely from noisy measurements.

To ensure smoothness and differentiability of \( U(t) \), we can construct it using cubic Hermite spline interpolation over discrete control points or other interpolation techniques. The Neural CDE model comprises three main components:
\begin{itemize}
    \item \textbf{Initial Encoder:} A neural network maps the initial observation \( X(t_0) \) (and optionally other context) into the initial latent state \( Z(t_0) \).
    \item \textbf{Vector Field:} The neural function \( f_\theta \) governs the evolution of \( Z(t) \) with respect to changes in \( U(t) \).
    \item \textbf{Decoder:} A linear projection maps each latent state \( Z(t_i) \) to an estimate of the time derivative \( \hat{X}'(t_i) \) of the observed signal.
\end{itemize}

The system is numerically integrated via the Diffrax python library~\cite{kidger2021on} using a differentiable explicit ODE solver~\cite{hairer2008solving-i} (for example, Dormand-Prince~\cite{dormand1980family, shampine1986some} or Tsitouras 5/4 Runge-Kutta method~\cite{tsitouras2011runge}) using adaptive time-stepping~\cite{hairer2002solving-ii, soderlind2002automatic}, enabling end-to-end training via discretise-then-optimise automatic differentiation~\cite{ma2021comparison, stumm2010new, wang2009minimal}. The learned derivatives serve as a smooth and robust approximation to the system's underlying dynamics, particularly beneficial for symbolic regression, which relies on clean gradient information.

\subsubsection*{State Trajectory Reconstruction via Integration}

While Neural CDEs are often used to directly predict state trajectories, we instead first estimate the time derivatives,
\[\hat{X}'(t_i) = \text{Decoder}(Z(t_i)).\]

Afterward, reconstruct the state \( \hat{X}(t) \) by numerical integration. For instance, applying a standard integrator or simple methods, such as the trapezoidal rule, we compute:
\[
\hat{X}(t_i)
\;\approx\;
X(t_0)
\;+\;
\sum_{j=0}^{i-1}
\frac{1}{2} \left( \hat{X}'(t_j) + \hat{X}'(t_{j+1}) \right)
\cdot
\left( t_{j+1} - t_j \right).
\]

This two-stage approach, derivative estimation followed by integration, provides a noise-robust estimate of the full state-trajectory. More importantly, the estimated derivatives \( \hat{X}'(t) \) can then be passed to symbolic regression algorithms to discover interpretable models.

\section{Experiments}
We demonstrate the effectiveness of our proposed KAN-SR framework on two different tasks integral to scientific discovery. First, we evaluate KAN-SR's performance on the SRSD-Feynmann~\cite{Matsubara2022} dataset, which is an adaptation of the widely adopted Feynmann equation dataset. In a second part, we extend the algorithm to the dynamic modelling of a bioprocess using neural-differential equations to extract the rate laws governing the process in a two-step approach.

It is important to mention here that all experiments were run without the additional Deeper-KAN loop (Figure~\ref{fig:deepkan}), to showcase the additional value of searching for simplifications. This allows us to circumvent using more complex models which may find solutions that are less interpretable.

\subsection{SRSD-Dataset}
The SRSD-Dataset~\cite{Matsubara2022} compared to the popular Feynman equation dataset, has more realistic sampling ranges of the independent variables and additional dummy variables. The improved sampling strategy makes the individual problems much more complex, as variables, depending on the set, can span many orders of magnitude, making the search much more difficult.\\ 
Contrary to standard machine learning methods, we are interested in finding not only low fitting error equations but also rediscovering the true underlying function. Our main objective is therefore the solution rate mentioned in La Cava et al.~\cite{Lacava2021} and not solely the accuracy metrics such as the mean squared error or the coefficient of determination.

Since in a scientific environment measurements are noisy, we also added different levels of white Gaussian noise to the dependent variable $y$ as a fraction of the signal root mean squared value.~\cite{Lacava2021}

\begin{align}
y_{\text{noise}} = y + \epsilon, \quad \epsilon \sim \mathcal{N} \left( 0, \gamma \sqrt{\frac{1}{N} \sum_{i=1}^N y_i^2} \right)
\label{eq:noise}
\end{align}

We compare our framework to other SR tools benchmarked in the SRSD paper~\cite{Matsubara2022}, namely gplearn, AFP, AFP-FE, AIF, DSR, E2E, uDSR and PySR~\cite{gplearn, Schmidt2011, Schmidt2009, Udrescu2020, Petersen2019, Kamienny2022, Landajuela2022, Cranmer2023}. For details of the baseline models, we refer the reader to the corresponding papers. For the implementation and hyperparameters of the models used in the benchmark, we refer to the original SRSD paper~\cite{Matsubara2022}.

\subsection{Solution Rate on the SRSD Dataset}
All methods benchmarked in this study, with the exception of KAN-SR, were evaluated in Matsubara et al.~\cite{Matsubara2022}.

\subsubsection*{Baseline results}
\begin{table}[H]
\centering
\small
\renewcommand{\arraystretch}{1.3}
\resizebox{\textwidth}{!}{
\begin{tabular}{l|ccccccccc}
\toprule
 \textbf{Group} & \textbf{gplearn} & \textbf{AFP} & \textbf{AFP-FE} & \textbf{AIF} & \textbf{DSR} & \textbf{E2E} & \textbf{uDSR} & \textbf{PySR} & \textbf{KAN-SR} \\
\midrule
Easy  & 6.67\% & 20.0\% & 23.3\% & 30.0\% & 46.7\% & 0.00\% & 50.0\% & 60.0\% & \cellcolor{gray!40}\textbf{93.3\%} \\
\hline
Medium & 0.00\% & 2.50\% & 2.50\% & 2.50\% & 10.0\% & 0.00\% & 17.5\% & 30.0\% & \cellcolor{gray!40}\textbf{60.0\%} \\
\hline
Hard   & 0.00\% & 0.00\% & 0.00\% & 2.00\% & 2.00\% & 0.00\% & 4.00\% & 4.00\% & \cellcolor{gray!40}\textbf{12.0\%} \\
\bottomrule
\end{tabular}
}
\vspace{0.5em}
\caption{Results of the solution rate~\cite{Lacava2021} of different methods on the SRSD-Feynman~\cite{Matsubara2022} dataset.}
\label{tab:baseline-results}
\end{table}

\subsubsection*{Results Including Dummy Variables}
\begin{table}[H]
\centering
\small
\renewcommand{\arraystretch}{1.3}
\resizebox{\textwidth}{!}{
\begin{tabular}{l|ccccccccc}
\toprule
\textbf{Group} & \textbf{gplearn} & \textbf{AFP} & \textbf{AFP-FE} & \textbf{AIF} & \textbf{DSR} & \textbf{E2E} & \textbf{uDSR} & \textbf{PySR} & \textbf{KAN-SR} \\
\midrule
Easy   & 0.00\% & 16.7\% & 16.7\% & 0.00\% & 10.0\% & 0.00\% & 10.0\% & 20.0\% & \cellcolor{gray!40}\textbf{86.7\%} \\
\hline
Medium & 0.00\% & 0.00\% & 0.00\% & 0.00\% & 0.00\% & 0.00\% & 7.50\% & 5.00\% & \cellcolor{gray!40}\textbf{55.0\%} \\
\hline
Hard   & 0.00\% & 0.00\% & 0.00\% & 0.00\% & 2.00\% & 0.00\% & 0.00\% & 0.00\% & \cellcolor{gray!40}\textbf{10.0\%} \\
\bottomrule
\end{tabular}
}
\vspace{0.5em}
\caption{Solution rate~\cite{Lacava2021} of different methods on the SRSD-Feynman dataset with dummy variables~\cite{Matsubara2022}.}
\label{tab:dummy-results}
\end{table}

\subsubsection*{Impact of Noise}
\begin{table}[h!]
\centering
\small
\renewcommand{\arraystretch}{1.3}
\resizebox{0.6\textwidth}{!}{
\begin{tabular}{l|ccc}
\toprule
\textbf{Noise-Level} & \textbf{Easy-Dataset} & \textbf{Medium-Dataset} & \textbf{Hard-Dataset} \\
\midrule
None   & 93.3\% & 60.0\% & 12.0\%  \\
\hline
0.001 & 80.0\% & 45.0\% & 8.00\% \\
\hline
0.01   & 76.7\% & 45.0\% & 8.00\% \\
\hline
0.1   & 53.3\% & 25.0\% & 4.00\% \\
\bottomrule
\end{tabular}
}
\vspace{0.5em}
\caption{Solution rate~\cite{Lacava2021} of KAN-SR on the SRSD-Feynman datasets~\cite{Matsubara2022} with varying noise levels according to Equation~\ref{eq:noise}.}
\label{tab:noise-results}
\end{table}

KAN-SR achieved the highest solution rates for all levels of difficulty in the SRSD-Feynman dataset, significantly outperforming previous methods (Table~\ref{tab:baseline-results}). In Easy problems, it solved 93.3\% of the tasks, compared to 60.0\% by the next-best method. Even on the Medium and Hard problems, where most other models struggled, KAN-SR reached 60.0\% and 12.0\% solution rates, respectively. When dummy variables were added to the equations, performance dropped for all methods, but KAN-SR remained far ahead, solving 86.7\% of Easy, 55.0\% of Medium, and 10.0\% of Hard problems (Table~\ref{tab:dummy-results}). \\
We also tested how well KAN-SR handled noise in the data. As expected, performance decreased as noise increased, but the model remained rather robust, solving over half of the Easy problems even with 0.1 noise (Table~\ref{tab:noise-results}). Overall, these results show that KAN-SR is not only accurate, but also robust with respect to irrelevant features and noisy data.

\subsection{In-Silico Bioprocess Model}

To evaluate our framework on dynamic systems, we benchmark it using an in-silico model of batch fermentation~\cite{turton2008}. The system is described by the following set of ordinary differential equations (ODEs):

\begin{align*}
\frac{dX}{dt} &= \mu \cdot X \\
\frac{dS}{dt} &= -\frac{1}{Y_{X/S}} \cdot \mu \cdot X \\
\frac{dP}{dt} &= \frac{Y_{P/S}}{Y_{X/S}} \cdot \mu \cdot X \\
\mu &= \mu_{\text{max}} \cdot \frac{S}{K_S + S} \cdot \frac{k_1(T)}{1 + k_2(T)} \cdot \left(1 - \frac{X}{K_\mu + X} \right)
\end{align*}

\(X\), \(S\), and \(P\) represent the concentrations of biomass, substrate, and product, respectively. The specific growth rate \(\mu\) is described using a Monod type model that accounts for substrate saturation, temperature effects on activation and inactivation, and inhibitory impact at high biomass levels.

The temperature dependence is modeled using Arrhenius-type equations:

\begin{align*}
k_1(T) &= A_1 \cdot \exp\left(-\frac{E_{A,1}}{R \cdot T}\right) \\
k_2(T) &= A_2 \cdot \exp\left(-\frac{E_{A,2}}{R \cdot T}\right)
\end{align*}

To recover the governing expressions that generated the state variables \(X\), \(S\), and \(P\), we follow a two-step procedure similar to Forster et al.~\cite{Forster2024}. First, a neural controlled differential equation (NCDE) model (see Section \ref{sec:NCDE}) \cite{chen2019, kidger2021on} is trained on observed time series data to obtain smooth derivatives. These estimated derivatives are then used as targets in a second symbolic regression step that aims to rediscover the underlying kinetic model.

Due to the high cost, typically associated with bioprocess experiments, the number of in-silico simulations was limited to 12. Each simulation spans 80 hours with uniform sampling every 2 hours, resulting in 40 samples per run. The initial conditions for biomass (\(X\)), substrate (\(S\)), and product (\(P\)) were sampled using a Latin hypercube design. The lower bounds were set to \([0.1,\ 50,\ 0]\) and the upper bounds to \([0.4,\ 90,\ 0.4]\). Since the simplified process does not contain any controlled variable, time is the only input to the control path in this case. Although a standard neural ODE would suffice without any controlled variables, we use the NCDE framework to support the direct extension to possibly more complex scenarios involving multiple controlled variables and showcase it here as a possible modeling framework for dynamical systems in the scope of symbolic regression.

\begin{table}[h!]
\centering
\begin{tabular}{lllc}
\hline
\textbf{Symbol} & \textbf{Value} & \textbf{Unit} & \textbf{Description} \\
\hline
$\mu_{\text{max}}$ & 0.25 & h$^{-1}$ & Maximum specific growth rate \\
$K_S$ & 105.4 & kg/m$^3$ & Monod saturation constant \\
$Y_{X/S}$ & 0.07 & dimensionless & Biomass yield on substrate \\
$Y_{P/S}$ & 0.167 & dimensionless & Product yield on substrate \\
$K_\mu$ & 121.87 & g/L & Inhibition constant for biomass \\
$T$ & 308.15 & K & Operating temperature (35°C) \\
$R$ & 0.0083145 & kJ/(K·mol) & Universal gas constant \\
$A_1$ & 130.03 & dimensionless & Pre-exponential factor (activation) \\
$E_{A,1}$ & 12.43 & kJ/mol & Activation energy \\
$A_2$ & 3.83$\cdot10^4$ & dimensionless & Pre-exponential factor (inactivation) \\
$E_{A,2}$ & 298.55 & kJ/mol & Inactivation energy \\
\hline
\end{tabular}
\vspace{0.5em}
\caption{Model parameters for the in-silico batch fermentation ODE system.}
\end{table}

With these parameters, the system reduces to:

\begin{align}
\frac{dX}{dt} &= \frac{30.87 \cdot X \cdot S}{(X + 121.87)(S + 105.4)}\label{eq:ode1} \\
\frac{dS}{dt} &= -\frac{440.99 \cdot X \cdot S}{(X + 121.87)(S + 105.4)}\label{eq:ode2} \\
\frac{dP}{dt} &= \frac{73.65 \cdot X \cdot S}{(X + 121.87)(S + 105.4)}\label{eq:ode3}
\end{align}

To test whether the NCDE model can extract smooth derivatives in the presence of observation noise, synthetic noise was added to the state variables using Equation~\ref{eq:noise} with \(\gamma=0.02\). Symbolic regression was then applied to the noisy derivative estimates.

\begin{figure}[H]
    \centering
    \includegraphics[width=\linewidth]{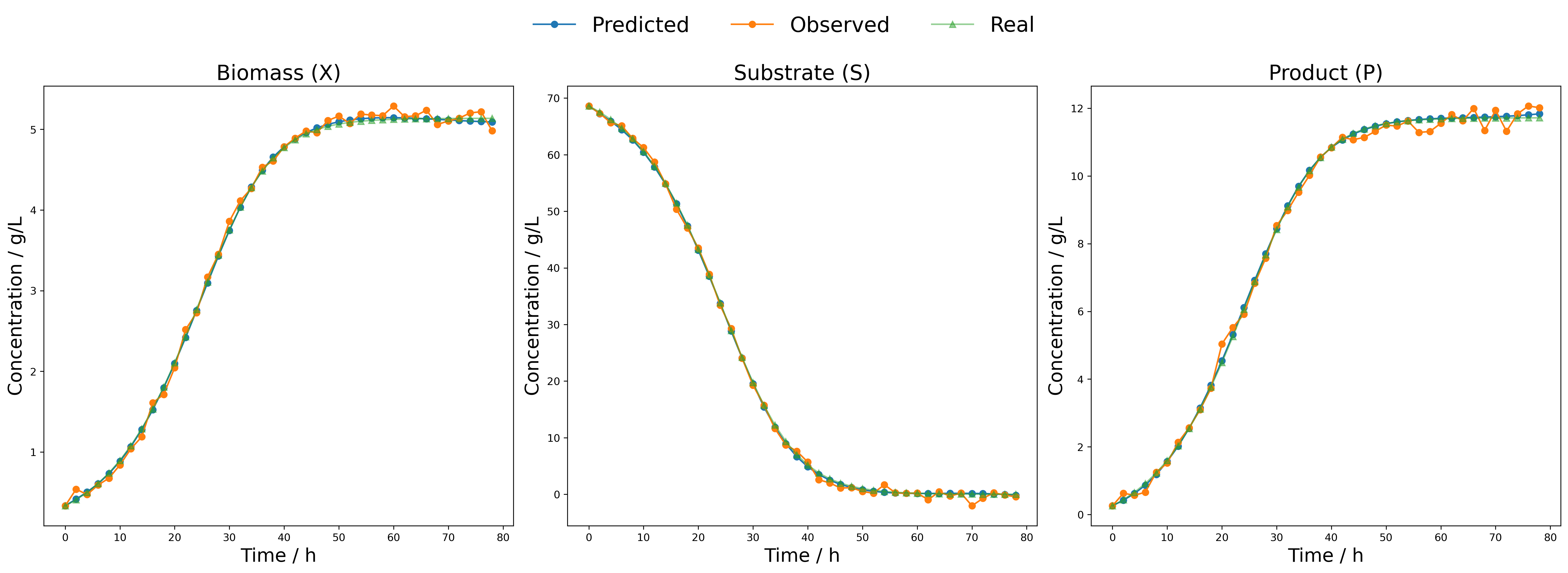}
    \caption{Sample concentration profiles of the NCDE model. The blue lines represent the predicted values by the NCDE model, the orange lines are the noisy, observed measurements and the green lines are the noise-free measurements.}
    \label{fig:sample-run}
\end{figure}

The trained NCDE model is able to model the process well, looking at Figure \ref{fig:sample-run} where a sample run is shown.

\begin{figure}[htbp]
    \centering
    \subfloat{
        \includegraphics[width=0.32\textwidth]{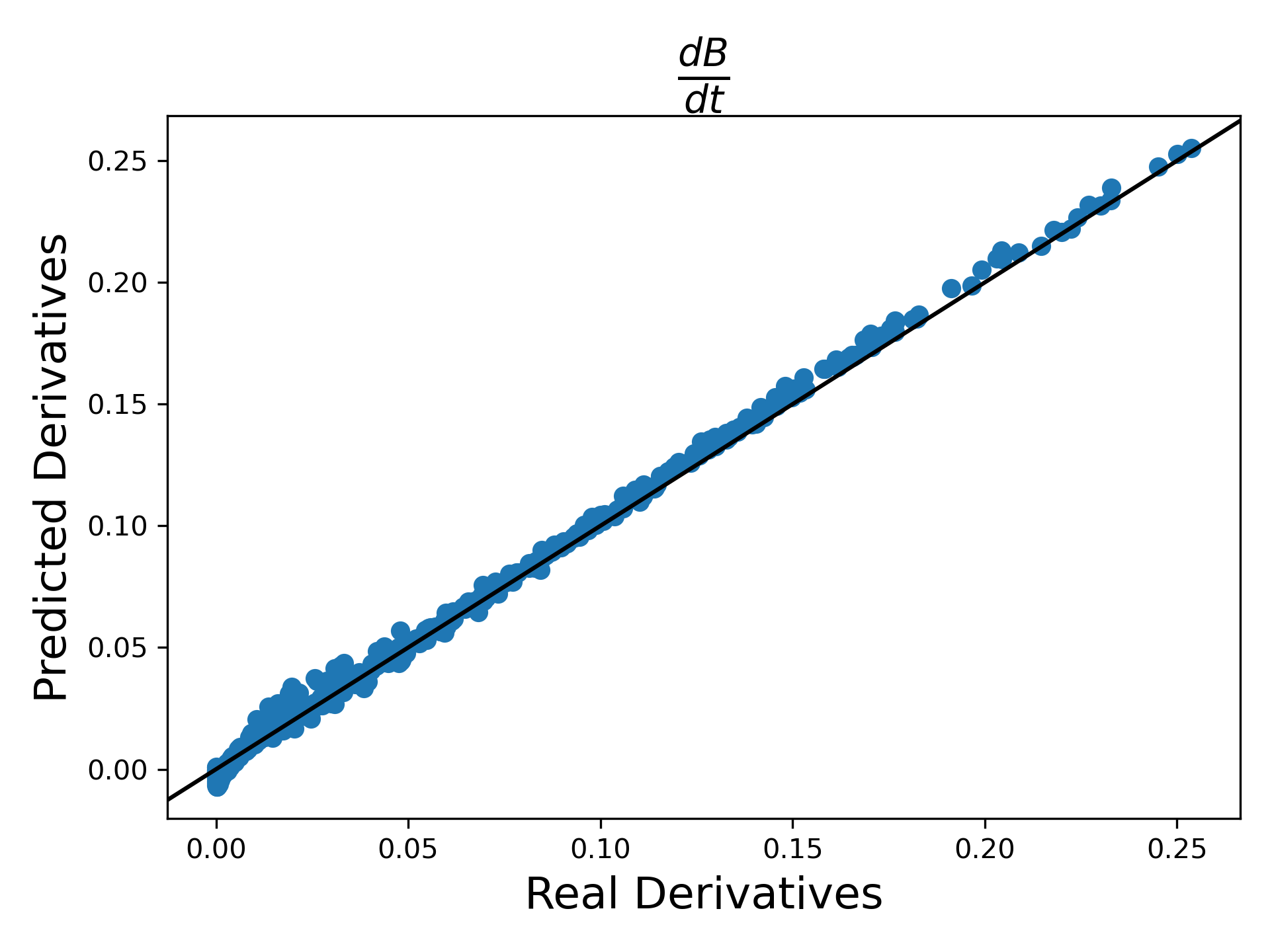}
    }
    \hfill
    \subfloat{
        \includegraphics[width=0.32\textwidth]{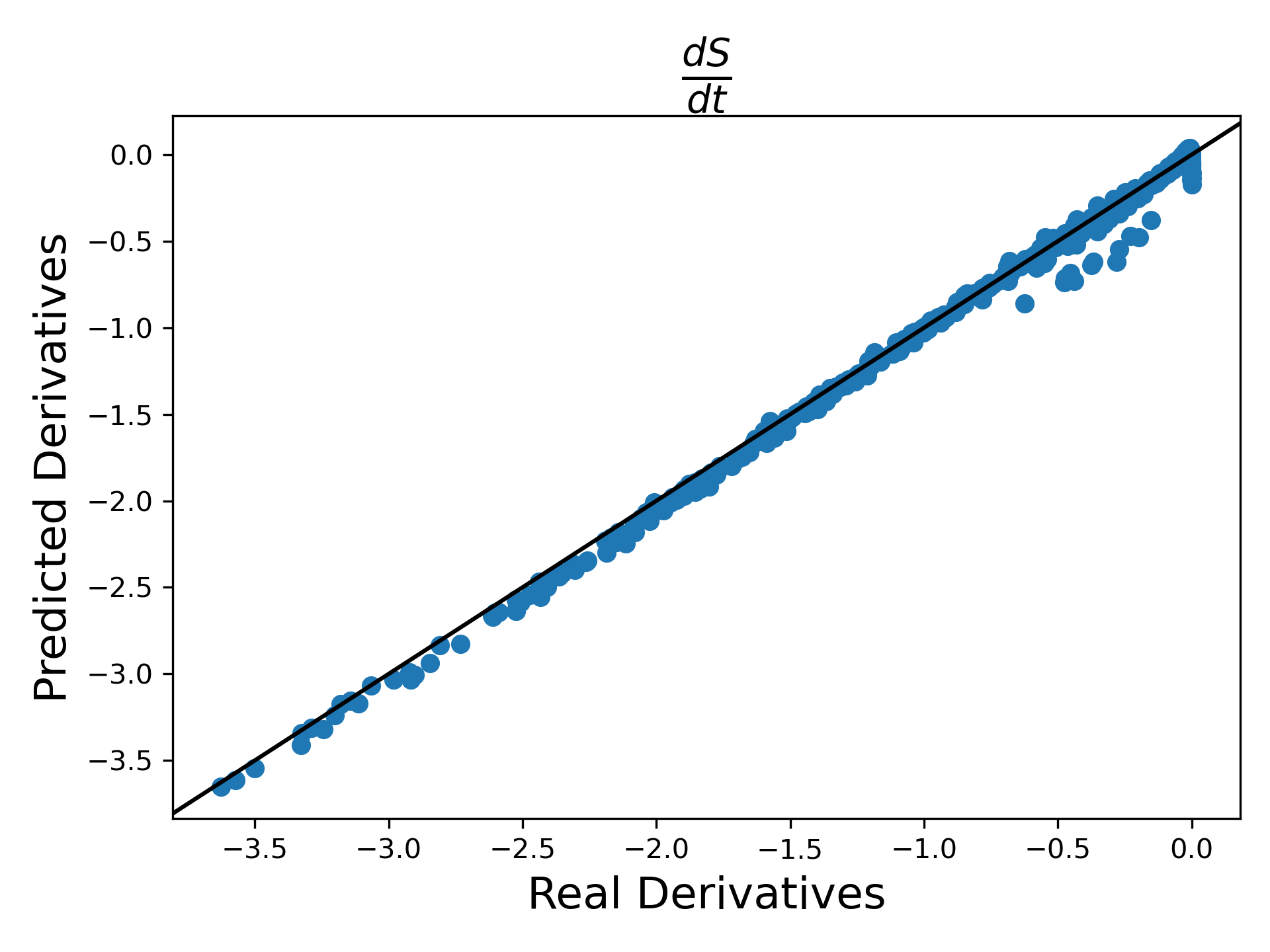}
    }
    \hfill
    \subfloat{
        \includegraphics[width=0.32\textwidth]{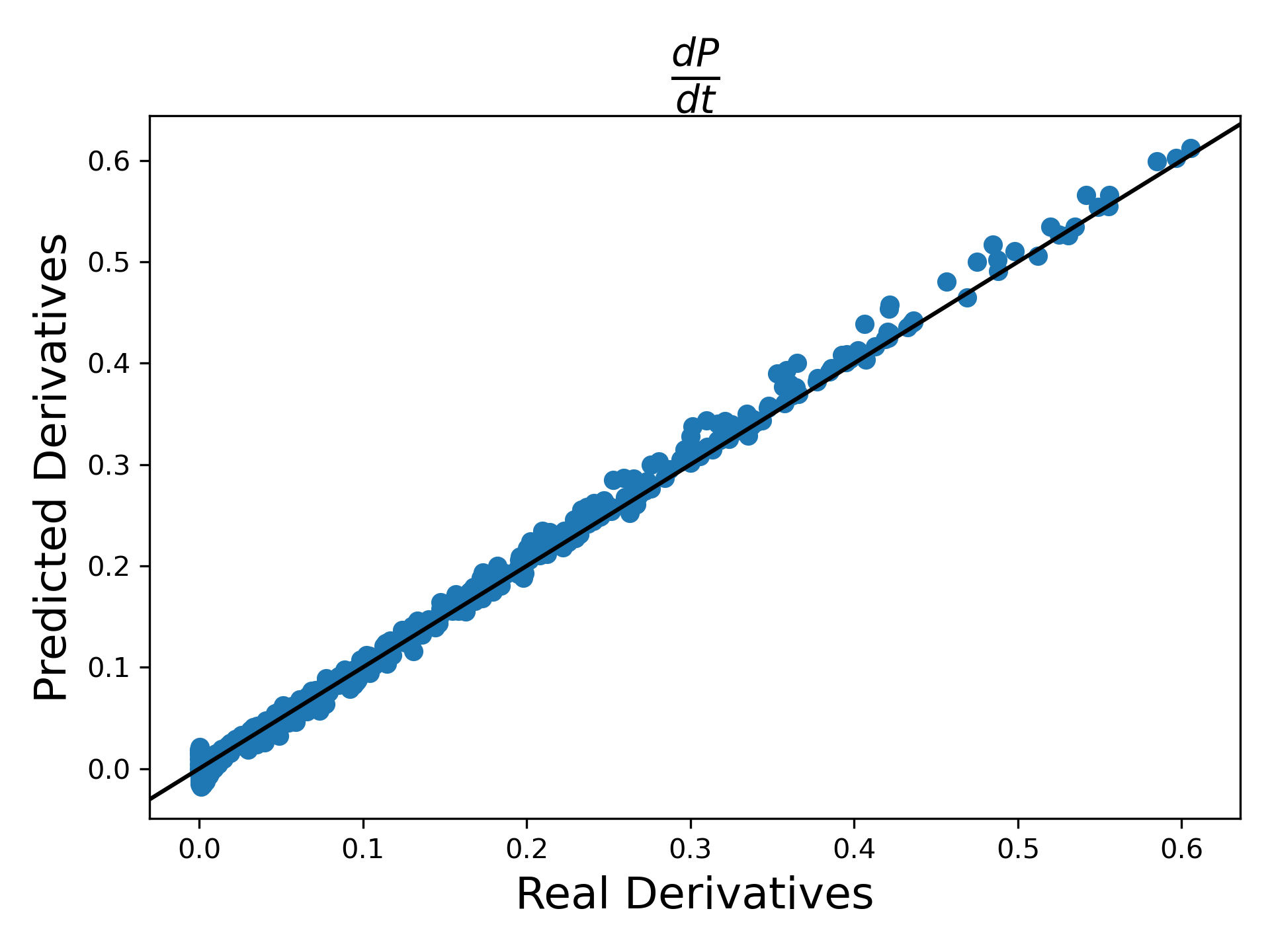}
    }
    \caption{Extracted derivatives of the NCDE model.}
    \label{fig:ncde_derivs}
\end{figure}

Looking at the parity plot of the extracted derivatives (Figure \ref{fig:ncde_derivs}), using the NCDE model, we are able to obtain smooth derivatives which are in good agreement with the real derivatives. 

These derivatives are then symbolically regressed using KAN-SR. The resulting system of ODEs, rounded to two decimal places, is:

\begin{align*}
 \frac{dX}{dt} &= \frac{0.25 \cdot X \cdot S}{S + 103.51} \\
 \frac{dS}{dt} &= -\frac{0.34 \cdot X \cdot S}{S + 118.76} \\
 \frac{dP}{dt} &= \frac{0.70 \cdot X \cdot S}{S + 129.51}
\end{align*}

Compared to Equations~\ref{eq:ode1}, \ref{eq:ode2}, and \ref{eq:ode3}, these expressions are structurally similar. However, the symbolic regression model does not recover the biomass-dependent denominator, suggesting that it did not identify the self-inhibition effect.

\begin{figure}[H]
    \centering
    \includegraphics[width=1\linewidth]{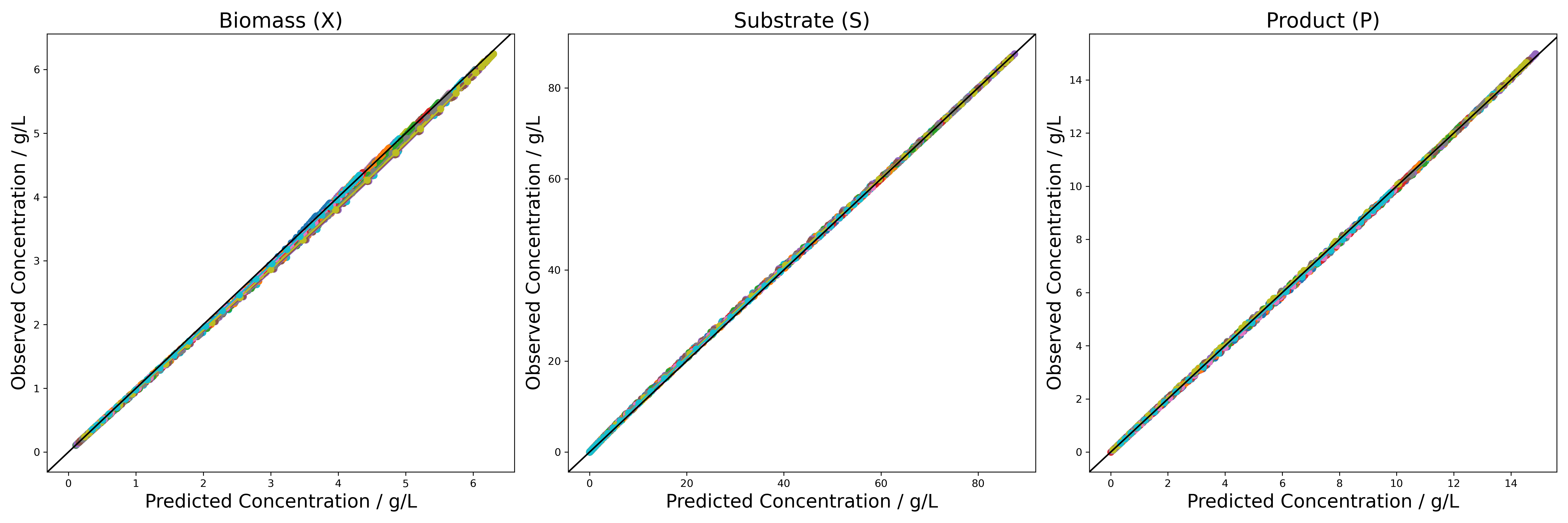}
    \caption{Parity plot of the different species concentrations after integrating the obtained symbolic derivatives.}
    \label{fig:pred_obs_symbolic_ode}
\end{figure}

Despite this, the parity plot in Figure~\ref{fig:pred_obs_symbolic_ode} shows that the model-generated state trajectories agree closely with observed test data, sampled within the same initial bounds. This indicates that the symbolic model can reproduce the system dynamics with only minor deviations.

\section{Conclusion}

Here, we presented KAN-SR as a symbolic regression framework which is capable of recovering mathematical expressions from noisy, high-dimensional data, as shown for numerical examples. Across the SRSD-Feynman benchmark, KAN-SR outperformed the investigated baseline methods in terms of solution rate, with notable gains if irrelevant variables were present in the data. This improvement can be attributed to the divide-and-conquer approach of combining sparse compositional univariate function learning and the simplification pipeline that assists in discovering simpler subproblems.

Combining additive and multiplicative KAN layers allows us to represent a wide range of expressions. Furthermore, the symbolic extraction stage, using a domain-informed library and nonlinear optimization, supports in recovering closed-form expressions that align with ground truth equations.

The extension to dynamic systems, as demonstrated in the bioprocess modeling case study, suggests that KAN-SR is capable of identifying governing equations in the context of neural differential equations. However, broader validation across real-world time-series systems is required to assess the generalizability and robustness of the framework under varying experimental conditions. Neural CDEs showed a promising way to integrate these dynamical systems into the KAN-SR framework, but it remains a two-step approach. A cleaner way would be to directly integrate symbolic regression algorithms into the Neural CDE, but this is out of scope for the proposed framework, due to the iterative manner of KAN-SR.

Although KAN-SR showed good performance across the SRSD-Feynman dataset, there are several limitations. The observed superior performance is only obtained when single-layer representations combined with the simplifications and the univariate library are enough to fit the equation. Our approach breaks down the symbolic regression problem, which is a large non-convex mixed integer non-linear programming problem in two levels. We restrict our search space to an upper bound of the number of operators using a fixed number of units in a single KAN layer and then further restrict the space by only fitting our limited univariate function library. This library by design may not contain all possible univariate functions that could describe our trainable activation function, and thus represents an inherent limitation.

Other algorithms such as GP explore the binary operator search space by evolving populations and performing non-linear constant fitting, while KAN-SR narrows the search space and may find local solutions. The drawback is that, if the optimal solution lies outside this constrained space, KAN-SR often fails to produce a good fitting equation, unlike other methods. Larger and deeper KANs can mitigate this, but excessive size makes them diffuse and hinders the extraction of symbolic expressions, whilst greatly increasing the number of tunable hyperparameters. In the studied physics case, all data points came from a closed-form equation, meaning the ground truth was recoverable from the training variables, unlike black-box or real-world problems where key variables might be missing. Although KAN-SR handles Gaussian noise well, this lack of complete information may prevent the discovery of good fitting equations.

\newpage
\bibliographystyle{unsrt}  
\bibliography{references}  
\end{document}